\documentclass[conference]{IEEEtran}
\IEEEoverridecommandlockouts
\usepackage{cite}
\usepackage{amsmath,amssymb,amsfonts}
\usepackage{algorithmic}
\usepackage{graphicx}
\usepackage{textcomp}
\usepackage{xcolor}
\def\BibTeX{{\rm B\kern-.05em{\sc i\kern-.025em b}\kern-.08em
    T\kern-.1667em\lower.7ex\hbox{E}\kern-.125emX}}

\usepackage{url}

\begin{document}

\title{Quantifying and Inducing Shape Bias in CNNs\\ via Max-Pool Dilation
}

\author{
    \IEEEauthorblockN{Takito Sawada, Akinori Iwata, and Masahiro Okuda}
    \IEEEauthorblockA{
        \textit{Doshisha University}, 
        Kyoto, Japan \\
        Emails: sawada@vig.doshisha.ac.jp, iwata@vig.doshisha.ac.jp, masokuda@mail.doshisha.ac.jp
    }
}

\maketitle

\begin{abstract}
Convolutional Neural Networks (CNNs) exhibit a well-known texture bias, prioritizing local patterns over global shapes---a tendency inherent to their convolutional architecture. While this bias is beneficial for texture-rich natural images, it often degrades performance on shape-dominant data such as illustrations and sketches.
Although prior work has proposed shape-biased models to mitigate this issue, these approaches lack a quantitative metric for identifying which datasets would actually benefit from such modifications.
To address this limitation, we propose a data-driven metric that quantifies the shape--texture balance within a dataset by computing the Structural Similarity Index (SSIM) between an image's luminance (Y) channel and its L0-smoothed counterpart.
Building on this metric, we introduce a computationally efficient adaptation method that promotes shape bias by modifying the dilation of max-pooling operations while keeping convolutional weights frozen. Experimental results demonstrate consistent accuracy improvements on shape-dominant datasets, particularly in low-data regimes where full fine-tuning is impractical, requiring training only the final classification layer.
\end{abstract}


\section{Introduction}
Convolutional Neural Networks (CNNs) exhibit a well-documented \textit{texture bias}, prioritizing local texture cues over global shape information \cite{geirhos2019imagenettexturebias}. While this bias is beneficial for natural images rich in textural detail, it often degrades performance on shape-dominant data such as illustrations and sketches, making bias mitigation a critical challenge.

Existing approaches to reducing texture bias can be broadly categorized into \textit{architectural modifications}---including increased dilation rates (expanding receptive fields) \cite{signals5040040} and large-kernel convolutions \cite{ding2022scalingkernels}---and \textit{data-driven methods}, such as shape-oriented data augmentation \cite{yoshihara2023does, 9856965}, debiasing strategies \cite{li2021shapetexturedebiasedneuralnetwork}, and the construction of shape-dominant training datasets \cite{published_papers/41553144, published_papers/45794689}. Although these methods have demonstrated promising performance gains, they share a fundamental limitation: the absence of a quantitative metric for determining whether a given dataset would actually benefit from shape-biased modeling. As a result, model selection and bias mitigation strategies remain largely heuristic and highly dataset-dependent.

To address this limitation, we propose a data-driven metric that quantifies the shape--texture balance within a dataset. Specifically, we compute the Structural Similarity Index (SSIM) between an image's luminance (Y) channel and its $L_0$-smoothed counterpart \cite{L0}. This metric leverages the property of $L_0$ gradient minimization to selectively suppress high-frequency texture components while preserving salient structural edges, thereby providing a principled measure of global shape dominance.

The main contributions of this work are summarized as follows:
\begin{itemize}
\item \textbf{L0-SSIM Metric:} We propose a novel quantitative metric based on $L_0$ gradient minimization to objectively assess the shape--texture balance of a dataset. The metric provides a reliable indicator of whether shape-biased modeling is appropriate.
\item \textbf{Metric--Performance Relationship:} Through experiments on six diverse small-scale datasets, we demonstrate a strong correspondence between L0-SSIM scores and the effectiveness of shape-biased models, validating the metric as a criterion for model selection.
\item \textbf{Efficient Adaptation via Max-Pooling Dilation:} We introduce a lightweight adaptation strategy that induces shape bias by dilating Max-Pooling layers while freezing all convolutional weights. This approach offers a computationally efficient alternative to full retraining and is particularly effective for datasets identified as shape-dominant by the proposed metric.
\end{itemize}

\section{Proposed Method}

\subsection{Quantifying Shape--Texture Balance via SSIM}
\label{sec:metric_quantification}
To quantify the shape--texture characteristics of a dataset, we first convert each image to the YCbCr color space and extract its luminance (Y) channel. We then apply $L_0$ gradient minimization smoothing \cite{L0} to generate a structure-preserved version of this channel, using a fixed smoothing parameter $\lambda = 0.01$ for all experiments. Based on these representations, the proposed metric is defined as the Structural Similarity Index (SSIM) computed between the original Y channel and its $L_0$-smoothed counterpart.

This metric exploits the property of $L_0$ smoothing to selectively suppress high-frequency texture components while preserving salient structural edges. Consequently, \textbf{texture-dominant images} are expected to exhibit substantial pixel-level discrepancies between the original and smoothed versions, leading to \textbf{lower SSIM scores}. In contrast, \textbf{shape-dominant images} largely retain their global structure after smoothing, resulting in \textbf{higher SSIM scores}. By averaging the SSIM values over all images in a dataset, we obtain a quantitative measure of its inherent shape--texture balance.

\subsection{Efficient Shape-Biased Model Adaptation via Max-Pool Dilation}
\label{sec:proposed_adaptation}
Building on the proposed metric, we introduce a computationally efficient strategy for adapting CNNs to shape-rich domains. Prior studies have shown that increasing the dilation of convolutional layers enlarges the Effective Receptive Field (ERF) and promotes shape bias \cite{signals5040040}; however, such modifications typically necessitate retraining a large number of learnable parameters.

To avoid this computational overhead, we freeze all convolutional weights of a pre-trained CNN and modify only the \texttt{dilation} parameter of its \textbf{max-pooling layers}, increasing it from \texttt{1} to \texttt{2}. Since max-pooling operations are parameter-free, this adjustment expands the ERF without disrupting the learned convolutional filters. We hypothesize that this architectural change encourages the aggregation of broader spatial context, thereby shifting the model's inductive bias toward global shape information.

In this configuration, we train \textbf{only the final classification layer} on the target dataset. This design enables effective shape-biased adaptation while substantially reducing computational costs and mitigating the risk of overfitting, which is particularly advantageous in low-data regimes.

\section{Experiments}
This section presents the experimental setup and results demonstrating the effectiveness of the proposed method.

\subsection{Datasets}
\label{sec:datasets}
\begin{figure}[tbp]
    \centering
    \setlength{\tabcolsep}{1pt}
    \begin{tabular}{ccc}
      \begin{minipage}[t]{0.32\linewidth}
        \centering
        \includegraphics[width=\linewidth, keepaspectratio]{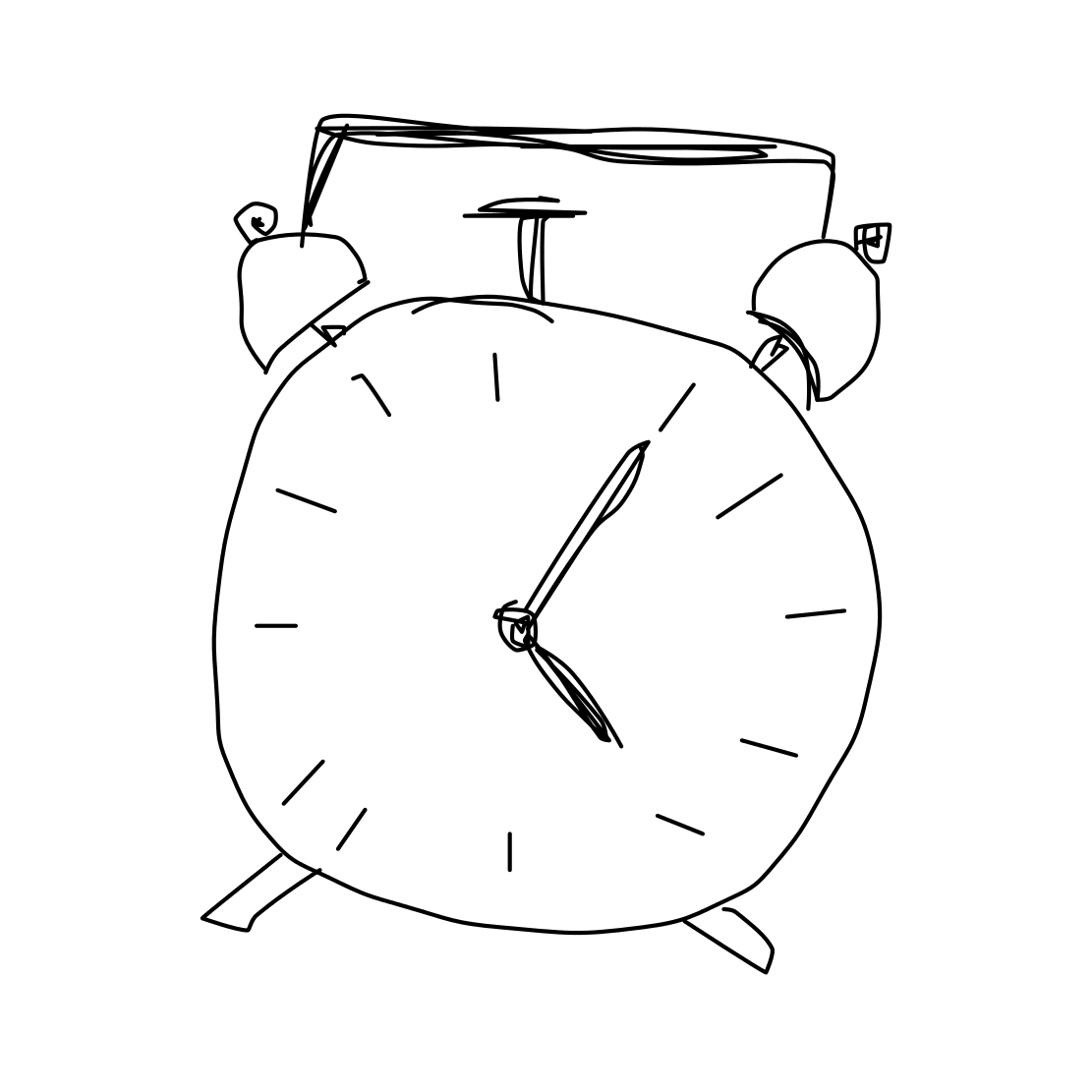}
        \centerline{(a) Sketches}
      \end{minipage} &
      \begin{minipage}[t]{0.32\linewidth}
        \centering
        \includegraphics[width=\linewidth, keepaspectratio]{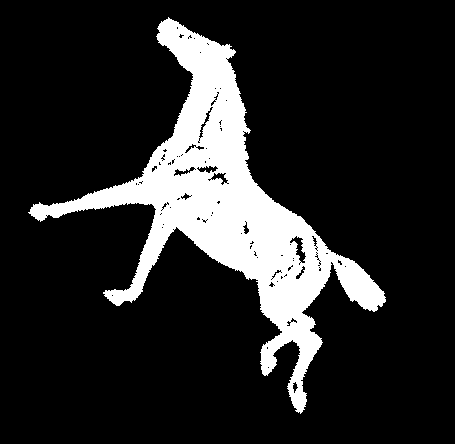}
        \centerline{(b) MPEG-7}
      \end{minipage} &
      \begin{minipage}[t]{0.32\linewidth}
        \centering
        \includegraphics[width=\linewidth, keepaspectratio]{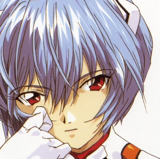}
        \centerline{(c) Anime}
      \end{minipage} \\
      
      \multicolumn{3}{c}{}\\[-1ex]

      \begin{minipage}[t]{0.32\linewidth}
        \centering
        \includegraphics[width=\linewidth, keepaspectratio]{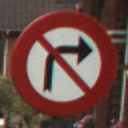}
        \centerline{(d) BTSD}
      \end{minipage} &
      \begin{minipage}[t]{0.32\linewidth}
        \centering
        \includegraphics[width=\linewidth, keepaspectratio]{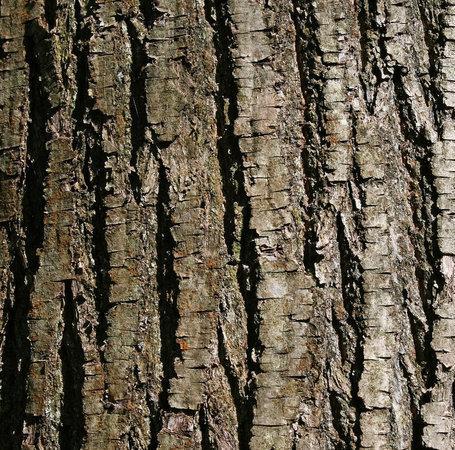}
        \centerline{(e) DTD}
      \end{minipage} &
      \begin{minipage}[t]{0.32\linewidth}
        \centering
        \includegraphics[width=\linewidth, keepaspectratio]{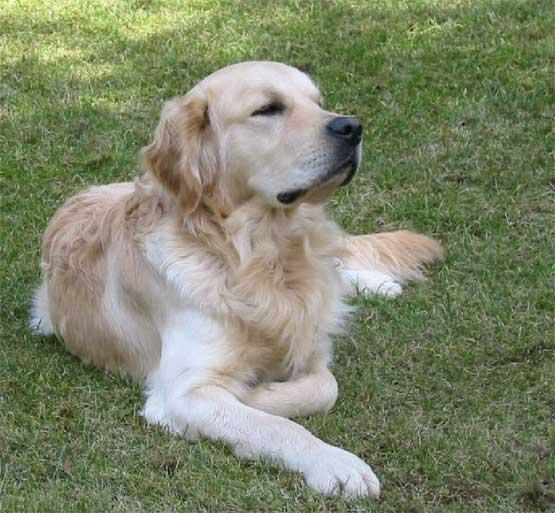}
        \centerline{(f) Dogs}
      \end{minipage}
    \end{tabular}
    \caption{Sample images from each dataset.}
    \label{fig:dataset_examples}
\end{figure}

To evaluate our method, we constructed six small-scale datasets from publicly available sources, each exhibiting distinct visual characteristics.
\textbf{TU-Berlin (Sketches)} \cite{eitz2012hdhso}, from which we randomly selected 100 classes (2,000 images) (Fig.~\ref{fig:dataset_examples}(a));
\textbf{MPEG-7} \cite{MPEG-7_dataset_url}, a binary silhouette dataset used in its entirety (70 classes, 1,400 images) (Fig.~\ref{fig:dataset_examples}(b));
\textbf{AnimeFace (Anime)} \cite{animeface_url}, from which we used 1,000 images from 50 randomly selected classes (Fig.~\ref{fig:dataset_examples}(c));
\textbf{BTSD} \cite{Timofte-WACV-2009}, from which we used 953 images (Fig.~\ref{fig:dataset_examples}(d));
\textbf{DTD} \cite{cimpoi14describing}, a texture dataset from which we randomly selected 940 images (Fig.~\ref{fig:dataset_examples}(e));
and \textbf{Stanford Dogs (Dogs)} \cite{KhoslaYaoJayadevaprakashFeiFei_FGVC2011}, a fine-grained dog breed dataset (a subset of ImageNet \cite{5206848}) from which we randomly selected 2,400 images (Fig.~\ref{fig:dataset_examples}(f)).

\subsection{Analysis via the Proposed Metric}
\label{sec:metric_analysis}

To validate the effectiveness of the proposed L0-SSIM metric (Section~\ref{sec:metric_quantification}), we computed the average L0-SSIM scores for the six datasets introduced in Section~\ref{sec:datasets}.
For each image, we calculated the SSIM between its luminance (Y) channel and the corresponding $L_0$-smoothed image~\cite{L0}, and then averaged the scores over the entire dataset.

The results are summarized in Table~\ref{tab:result1}.
As expected, datasets characterized by strong global shape information yielded high L0-SSIM scores, notably Sketches (0.999) and MPEG-7 (0.989).
In contrast, texture-dominant datasets such as DTD (0.709) and Stanford Dogs (0.699) exhibited substantially lower scores, indicating that $L_0$ smoothing induces significant structural changes by suppressing high-frequency texture details.
The Anime (0.821) and BTSD (0.810) datasets occupied an intermediate range between these two extremes.

Overall, these results demonstrate that the proposed metric effectively quantifies the degree of shape--texture bias inherent in each dataset.

\begin{table}[t]
    \caption{Average L0-SSIM Scores for Each Dataset}
    \label{tab:result1}
    \begin{center}
    \begin{tabular}{|l|c|}
        \hline
        \textbf{Dataset} & \textbf{Average L0-SSIM} \\
        \hline
        Sketches & 0.999 \\
        MPEG-7   & 0.989 \\
        Anime    & 0.821 \\
        BTSD     & 0.810 \\
        DTD      & 0.709 \\
        Dogs     & 0.699 \\
        \hline
    \end{tabular}
    \end{center}
\end{table}

\subsection{Experiment 1: Metric Validation}
\label{sec:metric_validation}

To examine whether the proposed L0-SSIM metric (Table~\ref{tab:result1}) serves as a reliable guide for model selection, we compared two models based on a pre-trained ResNeXt-50 architecture~\cite{xie2017aggregatedresidualtransformationsdeep}.
The first was a standard \textbf{Texture-Biased Model (T-Model)} with the default configuration (\texttt{dilation=1}).
The second was an \textbf{Existing Shape-Biased Model ($S_{\text{conv}}$-Model)}, in which, following prior work~\cite{signals5040040}, the \texttt{dilation} of all $3\times3$ convolutional layers was set to \texttt{3}.
For this model, we utilized the \textbf{official pre-trained weights provided by the authors of \cite{signals5040040}}, which were trained on ImageNet with the increased dilation.
For both models, all parameters were frozen except for the final classification layer, which was randomly initialized and trained.

The models were evaluated on the six small-scale datasets described in Section~\ref{sec:datasets} using 5-fold cross-validation, reporting the average classification accuracy.
We employed the Adam optimizer~\cite{kingma2014adam} with an initial learning rate of \texttt{1e-3}, decayed by a factor of \texttt{0.8} every \texttt{10} epochs using a StepLR schedule.
The batch size was set to \texttt{8}, with training performed for up to \texttt{100} epochs.
Early stopping was triggered if neither validation loss nor accuracy improved for \texttt{5} consecutive epochs (\texttt{patience=5}).
All input images were resized to $224\times224$ and normalized; Cross-Entropy Loss was used as the objective function.

The results in Table~\ref{tab:result2} exhibit a clear correspondence with the L0-SSIM scores reported in Table~\ref{tab:result1}.
For datasets identified as shape-dominant (high L0-SSIM), such as \textit{Sketches} (0.999) and \textit{MPEG-7} (0.989), the shape-biased $S_{\text{conv}}$-Model outperformed the T-Model by 3.1 and 2.1 percentage points (pp), respectively.
A similar performance gain was observed for intermediate datasets, including \textit{Anime} (0.821) and \textit{BTSD} (0.810).

In contrast, for datasets identified as texture-dominant (low L0-SSIM), this trend weakened or reversed.
On \textit{DTD} (0.709), the performance difference was marginal (+0.2 pp), whereas on \textit{Stanford Dogs} (0.699), the texture-biased T-Model outperformed the $S_{\text{conv}}$-Model by a substantial margin of 26.5 pp.

Overall, these results demonstrate that the proposed L0-SSIM metric effectively captures dataset-level characteristics and provides a reliable criterion for selecting models with an appropriate inductive bias.

\begin{table}[t]
  \caption{Classification Accuracy (\%) of Models with Different Convolutional Dilation Rates}
  \label{tab:result2}
  \begin{center}
  \setlength{\tabcolsep}{3pt}
  \begin{tabular}{|l|c|c|c|c|c|c|}
    \hline
    \textbf{Model} & \textbf{Sketches} & \textbf{MPEG-7} & \textbf{Anime} & \textbf{BTSD} & \textbf{DTD} & \textbf{Dogs} \\
    \hline
    T & 54.5 & 92.9 & 70.8 & 84.7 & 46.6 & \textbf{71.5} \\
    \hline
    $S_{\text{conv}}$ & \textbf{57.6} & \textbf{95.0} & \textbf{75.1} & \textbf{85.8} & \textbf{46.8} & 45.0 \\
    \hline
  \end{tabular}
  \end{center}
\end{table}

\subsection{Experiment 2: Adaptation Method Validation}
\label{sec:proposed_method_validation}

Next, we evaluated our second proposal (Section~\ref{sec:proposed_adaptation}): the \textit{weight-frozen, max-pooling-dilated model} ($S_{\text{maxpool}}$-Model).
In this model, the \texttt{dilation} of all max-pooling layers is set to \texttt{2}, while all convolutional weights remain frozen.
We compared its performance against the baseline T-Model (\texttt{dilation=1}) under identical experimental conditions to those in Experiment~1, including 5-fold cross-validation.

The results are summarized in Table~\ref{tab:result3}.
The $S_{\text{maxpool}}$-Model exhibited a clear shift toward shape bias, achieving accuracy improvements on shape-oriented datasets such as \textit{MPEG-7} (+0.5 pp) and \textit{BTSD} (+1.7 pp).
In contrast, performance slightly degraded on the highly abstract \textit{Sketches} dataset ($-0.9$ pp).
Consistent with this trend, performance on texture-dominant datasets declined, most notably on \textit{Stanford Dogs} ($-12.8$ pp).

Overall, these results demonstrate that dilating max-pooling layers provides a \textbf{computationally efficient mechanism for inducing shape bias} without retraining convolutional weights.
Although this configuration is less effective for highly abstract imagery, the observed performance degradation on texture-rich datasets reflects an expected trade-off, indicating a reduced reliance on texture cues rather than a methodological limitation.

\begin{table}[t]
  \caption{Classification Accuracy (\%) of Models with Different Max-Pooling Dilation Rates}
  \label{tab:result3}
  \begin{center}
  \setlength{\tabcolsep}{3pt}
  \begin{tabular}{|l|c|c|c|c|c|c|}
    \hline
    \textbf{Model} & \textbf{Sketches} & \textbf{MPEG-7} & \textbf{Anime} & \textbf{BTSD} & \textbf{DTD} & \textbf{Dogs} \\
    \hline
    T & \textbf{54.5} & 92.9 & \textbf{70.8} & 84.7 & \textbf{46.6} & \textbf{71.5} \\
    \hline
    $S_{\text{maxpool}}$ & 53.6 & \textbf{93.4} & 69.2 & \textbf{86.4} & 42.9 & 58.7 \\
    \hline
  \end{tabular}
  \end{center}
\end{table}

\section{Discussion}
\label{sec:discussion}
Our experiments validate the proposed L0-SSIM metric as a quantitative indicator of dataset bias and clarify the efficacy and limitations of the Max-Pooling dilation–based adaptation method.

\subsection{Analysis and Practical Implications}

\paragraph{L0-SSIM as an Indicator of Dataset Bias}
Results from Experiment~1 (Table~\ref{tab:result2}) demonstrate that the L0-SSIM metric (Table~\ref{tab:result1}) reliably predicts the optimal inductive bias for a given dataset.
For datasets with high L0-SSIM scores, identified as shape-dominant (e.g., \textit{Sketches} and \textit{MPEG-7}), the shape-biased $S_{\text{conv}}$-Model consistently outperformed the texture-biased T-Model.
Conversely, for datasets with low L0-SSIM scores, identified as texture-dominant (e.g., \textit{DTD} and \textit{Dogs}), the T-Model was competitive or markedly superior.
Notably, the 26.5 percentage-point performance drop observed when applying the $S_{\text{conv}}$-Model to the \textit{Dogs} dataset confirms that enforcing a strong shape bias on texture-rich data is counterproductive.
Overall, these results indicate that L0-SSIM effectively captures the shape–texture characteristics of datasets and serves as a reliable criterion for model bias selection.

\paragraph{Efficacy and Trade-offs of Max-Pooling Dilation}
Experiment~2 (Table~\ref{tab:result3}) evaluated the proposed $S_{\text{maxpool}}$-Model and demonstrated that Max-Pooling dilation provides a computationally efficient mechanism for inducing shape bias, albeit with dataset-dependent efficacy.
By modifying only the non-learnable \texttt{dilation} parameter of Max-Pooling layers while freezing all convolutional weights, the model achieved improved accuracy on shape-oriented datasets such as \textit{BTSD} ($+1.7$~pp) and \textit{MPEG-7} ($+0.5$~pp).
However, it showed no improvement on the highly abstract \textit{Sketches} dataset ($-0.9$~pp) and caused substantial degradation on texture-dominant datasets such as \textit{Dogs} ($-12.8$~pp).
This behavior reflects an inherent trade-off: strengthening shape bias inevitably reduces sensitivity to local texture cues.
Nevertheless, the simplicity and efficiency of Max-Pooling dilation make it an attractive adaptation mechanism in practice.

A key advantage of the proposed method lies in its computational and energy efficiency.
Unlike full fine-tuning, which requires costly backpropagation through millions of learnable parameters, our weight-frozen adaptation modifies only the non-learnable dilation parameters of Max-Pooling layers.
As a result, gradient computation in convolutional layers is completely avoided, incurring negligible additional computational and energy overhead during adaptation.
Consequently, the proposed framework is particularly well suited for deployment in resource-constrained or low-power environments.

\paragraph{Metric-Guided Efficient Model Adaptation}
Taken together, these findings suggest that the $S_{\text{maxpool}}$-Model should not be applied indiscriminately, but rather under the guidance of the L0-SSIM metric.
We therefore propose a two-stage framework: first, assess the dataset’s shape–texture balance using L0-SSIM; second, apply the computationally efficient $S_{\text{maxpool}}$-Model \textit{only} when the dataset is identified as shape-dominant (e.g., L0-SSIM $> 0.8$).
This metric-guided strategy prevents unnecessary performance degradation on texture-rich datasets while enabling effective bias adaptation for shape-oriented data, particularly in data-scarce scenarios where full fine-tuning is impractical.
In practical inference scenarios, test samples often arrive sequentially rather than as a static batch.
In such cases, the L0-SSIM metric can be estimated incrementally by maintaining a running average over the incoming stream, enabling bias assessment without requiring access to the full dataset.
This capability allows the proposed framework to be naturally extended to online or streaming inference scenarios.

\subsection{Limitations and Future Work}
While the proposed framework demonstrates efficacy on small-scale datasets using CNN architectures, several limitations and directions for future work remain.
First, the applicability of Max-Pooling dilation to Transformer-based models, which capture shape and texture cues through fundamentally different mechanisms, requires further investigation.
Second, although the shape-dominance threshold was empirically set to 0.8 based on the representative datasets in this study, a more generalized determination—such as through ROC analysis on larger-scale datasets—would further strengthen the metric’s reliability.
Finally, the sensitivity of the L0-SSIM metric to common data augmentations, including color jitter, cropping, and noise, remains to be systematically quantified.
Evaluating and enhancing the robustness of the metric under such variations is a critical direction for future work, particularly to ensure stable deployment in real-world scenarios.

\section{Conclusion}
\label{sec:conclusion}
This paper addressed the challenge of adapting pre-trained CNNs whose inherent texture bias may not align with the characteristics of small-scale datasets.
To this end, we proposed a two-component framework consisting of a quantitative dataset-bias metric and a computationally efficient, weight-frozen adaptation method.
First, we introduced the \textbf{L0-SSIM metric}, which quantifies the shape–texture balance of a dataset by computing the SSIM between an image’s luminance (Y) channel and its L0-smoothed counterpart.
Experimental results (Table~\ref{tab:result1}) demonstrated that this metric reliably predicts when a shape-biased model is likely to outperform a texture-biased baseline (Table~\ref{tab:result2}).
Second, we proposed a lightweight \textbf{Max-Pooling Dilation} adaptation strategy that modifies only the non-learnable dilation parameters while keeping all convolutional weights frozen.
This approach improved performance on shape-dominant datasets (e.g., \textit{BTSD} and \textit{MPEG-7}), while degrading accuracy on texture-rich (e.g., \textit{Dogs}) or highly abstract (e.g., \textit{Sketches}) datasets (Table~\ref{tab:result3}), reflecting the expected trade-off induced by stronger shape bias.
The primary contribution of this work lies in integrating these components into a practical, \textbf{metric-guided adaptation framework}.
We conclude that L0-SSIM should first be used to assess dataset bias; when a dataset is identified as shape-dominant (e.g., L0-SSIM~$>$~0.8), the proposed adaptation method provides an effective and low-cost means of bias alignment without retraining convolutional weights.
Future work will explore a broader range of dilation configurations and investigate automated parameter selection driven by the L0-SSIM score.

\vspace{12pt}

\end{document}